\documentclass{article} % For LaTeX2e
\usepackage{iclr2023_conference_tinypaper,times}

\usepackage{amsmath,amsfonts,bm}

\def\eqref#1{equation~\ref{#1}}
\def\1{\bm{1}}

\DeclareMathAlphabet{\mathsfit}{\encodingdefault}{\sfdefault}{m}{sl}
\SetMathAlphabet{\mathsfit}{bold}{\encodingdefault}{\sfdefault}{bx}{n}

\usepackage{graphicx}
\usepackage{hyperref}
\usepackage{url}

\usepackage{verbatim}

\usepackage{xcolor}
\definecolor{mygray}{gray}{0.9}
\usepackage{fvextra}
\usepackage{tcolorbox}

\hypersetup{colorlinks = true, urlcolor = blue, linkcolor = blue, citecolor=blue}

\newenvironment{code}%
 {\VerbatimEnvironment
  \begin{tcolorbox}[colback=white, boxsep=0pt, arc=0pt, boxrule=0pt, frame empty]
  \begin{Verbatim}[fontsize=\small, commandchars=\\\{\},
    breaklines, breakafter=*, breaksymbolsep=0.5em,
    breakaftersymbolpre={\,\ensuremath{\rfloor}}]}%
 {\end{Verbatim}%
  \end{tcolorbox}}

\title{Neural Controlled Differential Equations with quantum hidden evolutions}

\author{Lingyi Yang \& Zhen Shao  \\
Mathematical Institute\\
University of Oxford\\
\texttt{\{yangl, shaoz\}@maths.ox.ac.uk} \\
}
\iclrfinalcopy % Uncomment for camera-ready version, but NOT for submission.
\begin{document}

\maketitle
\vspace{-1em}

\begin{abstract}
We introduce a class of neural controlled differential equation inspired by quantum mechanics. 
Neural quantum controlled differential equations (NQDEs) model the dynamics by analogue of the Schr\"{o}dinger equation.
Specifically, the hidden state represents the wave function, and its collapse leads to an interpretation of the classification probability. We implement and compare the results of four variants of NQDEs on a toy spiral classification problem.
\end{abstract}

\vspace{-0.6em}

\section{Introduction}

\vspace{-0.15em}

Controlled differential equations (CDEs) model the dynamics of a sequential output $Y_t\in\mathbb{R}^p$ in response to a sequential input $X_t\in\mathbb{R}^q$ by 
\begin{equation}\label{eqn:cde}
    dY_t = f(Y_t, t) dX_t,
\end{equation}
for some fixed vector field $f(Y_t, t)$. Neural controlled differential equations (NCDEs) introduced by \citet{kidger2021neural} learn a CDE for a hidden variable \(z_t\).
The vector field 
$f(z_t, t)$ is modelled by a neural network $f(z_t, t; \theta)$ where $\theta$ are learnable parameters. The output is then \(Y_t = l(z_t)\), for some linear function $l$. The parameters \(\theta\) are fitted by the given input sequences $X_t$ and the outputs $Y_t$. 

Probabilistic models are popular for alleviating the issue of overfitting, motivating heuristics such as dropout. 
Quantum mechanics have probabilistic interpretation and is key in modelling physical phenomena.
We introduce architectures inspired by the Schr\"{o}dinger equation to model the latent space of NCDEs. Analogous to the Born interpretation for the collapse of the wave function, we have a \textit{collapse} function for observation times (where we need to make an inference for \(Y_t\)). As quantum systems and unitary systems (where the transition is unitary i.e. $U^*U =I$) are intricately linked, we implement and apply four variants to a toy spiral classification problem.

\vspace{-0.4em}

\section{Neural Quantum Differential Equations}

\vspace{-0.15em}

Complex numbers are indispensable for quantum mechanics. Complex recurrent neural networks, e.g. uRNN \citep{arjovsky16unitary} and ceRNN \citep{shafran2019complex}, have been studied to derive stability and convergence results. 
\citet{barrachina2023theory} provides a Python library for complex neural networks complete with backpropogation and other real neural network equivalents like max-pooling for complex functions. 
For other related literature, see Appendix~\ref{app:related_work}.

Combining neural controlled differential equations with quantum concepts has not yet been explored. 
We introduce a family of models based on quantum mechanical postulates. 
In particular, in the quantum world, the state of a system is represented by a complex wave function $\psi(x, t)$. 
Measurements/observables are modelled as linear operators on $\psi(x, t)$. 
The set of possible outcomes of these measurements are eigenvalues of this operator. 
The probability of obtaining any particular eigenvalue as the observation is proportional to the inner product between its associated eigenvector and the state $\psi(x, t)$. 
Therefore, the normalised inner products represent a probability distribution over the states.
The hidden quantum state evolves according to the Schr\"{o}dinger equation \citep{dirac1981principles}
\begin{equation}
    d\psi(x, t) = -i\hbar H\psi(x, t) dt,
\end{equation}
where $\hbar$ is the normalised Planck constant and $H$ is the Hamiltonian.
The evolution of the quantum state is a unitary operator, and the exponential of a time-independent Hamiltonian generates this unitary operator.

Our model is inspired by the practical success of neural differential equation based models, and the success of the quantum physics postulate to model physical systems. 
To obtain a neural quantum controlled differential equation (NQDE), we suppose that the dynamics of the latent state \(z_t\) is modelled by the Schr\"{o}dinger equation, driven by some control path \(X_t\), that is, 
\begin{equation}
    dz_t = -i\hbar Hz_t dX_t.
\end{equation}
Note that \(-i\hbar Hz_t\) is playing the role of the vector field \(f\) in the CDE (\ref{eqn:cde}).
For this complex vector field, \(-i\hbar Hz_t\), unitary conditions are imposed.
Unitary matrices are known to have good stability properties. ExpRNNs \citep{lezcano2019cheap} ensure orthogonality/unitarity using the exponential map from Lie group theory. The projUNN method developed by \citet{kiani2022projunn} projects the updated matrix back into the class of unitary matrices.

For each time that we need to ``observe'' the hidden state and make an inference for the output \(Y_t\), the hidden states pass through an operation that we refer to as the \textit{collapse}. For a classification problem with \(m\) classes, the collapse function is given by $g: \mathbb{C}^m \to \mathbb{R}^m$ or equivalently $\tilde{g}: \mathbb{R}^{2m} \to \mathbb{R}^m$, where \(\tilde{g}\) is composed of $g_1: \mathbb{R}^{2m} \to \mathbb{R}^{m}$, $g_2: \mathbb{R}^{m} \to \mathbb{R}^{m}$, and $g_3: \mathbb{R}^{m} \to \mathbb{R}^{m}$. The function $g_1$ takes the squared modulus of the complex input (represented in the real space). Then $g_2$ normalises the output from \(g_1\) to have norm 1, thus the output of \(g_2\) is a probability distribution, which we can then sample ($g_3$) for the output \(Y_t\). Therefore 
\[Y_t = \tilde{g}(z_t) = g_3(g_2(g_1(z_t))).\] 
This is analogous to the quantum system collapsing to an eigenstate.
We see that in practice a softmax can be utilised for \(g_2\).

\vspace{-0.3em}

\section{Experimental results}

\vspace{-0.1em}

We look a classification problem on the bi-directional spiral dataset as detailed by \citet{NEURIPS2018_69386f6b} using 128 spirals. 
We implemented four vector fields. Two of these have unitary constraints as imposed by ProjUNN \citep{kiani2022projunn} (and denoted by \_unn in the name) and the other two with orthogonal constraints using GeoTorch \citep{lezcano2019trivializations} (denoted by \_geo in the name). 
For each constraint method, we look at two variants: the first looks at modelling each class of the classification task separately, then concatenates the results together before a final linear layer (NQDE1\_unn and NQDE3\_geo) and the second performs the concatenation after the linear layer (NQDE2\_unn and NQDE4\_geo).
For specific architectures that we utilised and the number of trainable parameters, see Appendix~\ref{app:architecture}.
The results are given in Table~\ref{tab:results}. Hyperparameters can be found in Appendix~\ref{app: hyperparams}. 

\begin{table}[h]
\begin{tabular}{l|l|l|l|l}
\hline
\hline
\textbf{model}            & \textbf{final loss}      & \textbf{forward NFE} & \textbf{backward NFE} &  \textbf{accuracy}   \vspace{0.2em}   \\

\hline
\hline
NQDE1\_unn      & \textbf{0.00028 (0.0004)} & \textbf{1069.79 (58.71)}     & \textbf{2337.67 (431.67)}    &{1.000 (0) }    \\
NQDE2\_unn      & 0.00717 (0.0111)  & 1102.86 (54.95)    & {3093.38 (489.91) }    & 1.000 (0) \\
NQDE3\_geo & {0.12472 (0.2095)} & 1348.19 (43.75)    & 6781.65 (1690.76)     & {1.000 (0) }      \\
NQDE4\_geo & 0.03786 (0.0167) & 1288.21 (325.52)     & 4425.68 (1561.47)      & 1.000 (0)
\end{tabular}\caption{Spiral classification results on various architectures. Standard deviation in brackets are reported over 3 repeats.}\label{tab:results}
\end{table}

The models all use 20 epochs so that we can compare data efficiency. Given very limited data of only 128 spirals, we see that all of these architectures learn relevant dynamics for spiral classification and can reach 100\% accuracy after hyperparameter optimisation. 
Using orthogonal linear layers with GeoTorch 
requires more function evaluations. 
This is not unexpected as ProjUNN architectures makes a rank-$k$ approximation for computational efficiency. 
Using ProjUNN with the concatenation occurring before the linear layer gives the best model in terms of both loss and has the smallest number of function evaluations (NFEs).
The code for the experiments can be found at the Github repository \href{https://github.com/lingyiyang/NQDE}{https://github.com/lingyiyang/NQDE}.

\vspace{-0.3em}

\section{Conclusion/Discussion}

\vspace{-0.1em}

We have demonstrated that neural controlled differential equation architectures that emulate quantum evolutions can learn relevant dynamics on a toy classification problem. 
For future work, we would like to explore the approximation power of these models in greater depths (to derive similar results to \citet{VOIGTLAENDER202333}) as well as compare with other models on larger datasets.


\subsubsection*{Acknowledgements}
L.Y. was supported by the Alan Turing Institute, EPSRC [EP/S026347/1], and the Hong Kong Innovation and Technology Commission (InnoHK Project CIMDA). Z. S. was supported by the EPSRC [EP/S026347/1].

\subsubsection*{URM Statement}
The authors acknowledge that at least one key author of this work meets the URM criteria of ICLR 2024 Tiny Papers Track.
% Please include this URM Statement section at the end of the paper but before the references before. In your anonymized submission, we recommend stating ``The authors acknowledge that at least one key author of this work meets the URM criteria of ICLR 2024 Tiny Papers Track.'' For the camera ready version, we ask authors to identify which author(s) meet the URM criteria, e.g., ``Author TFB meets the URM criteria of ICLR 2024 Tiny Papers Track.'' The authors are also welcome to come up with their own phrases to affirm meeting this criteria.

\bibliography{ref}

\begin{thebibliography}{30}
\providecommand{\natexlab}[1]{#1}
\providecommand{\url}[1]{\texttt{#1}}
\expandafter\ifx\csname urlstyle\endcsname\relax
  \providecommand{\doi}[1]{doi: #1}\else
  \providecommand{\doi}{doi: \begingroup \urlstyle{rm}\Url}\fi

\bibitem[Arjovsky et~al.(2016)Arjovsky, Shah, and Bengio]{arjovsky16unitary}
Martin Arjovsky, Amar Shah, and Yoshua Bengio.
\newblock Unitary evolution recurrent neural networks.
\newblock In Maria~Florina Balcan and Kilian~Q. Weinberger (eds.), \emph{Proceedings of The 33rd International Conference on Machine Learning}, volume~48 of \emph{Proceedings of Machine Learning Research}, pp.\  1120--1128, New York, New York, USA, 20--22 Jun 2016. PMLR.

\bibitem[Barrachina et~al.(2023)Barrachina, Ren, Vieillard, Morisseau, and Ovarlez]{barrachina2023theory}
Jose~Agustin Barrachina, Chengfang Ren, Gilles Vieillard, Christele Morisseau, and Jean-Philippe Ovarlez.
\newblock Theory and implementation of complex-valued neural networks.
\newblock \emph{arXiv preprint arXiv:2302.08286}, 2023.

\bibitem[Chen et~al.(2018)Chen, Rubanova, Bettencourt, and Duvenaud]{NEURIPS2018_69386f6b}
Ricky T.~Q. Chen, Yulia Rubanova, Jesse Bettencourt, and David~K Duvenaud.
\newblock Neural ordinary differential equations.
\newblock In S.~Bengio, H.~Wallach, H.~Larochelle, K.~Grauman, N.~Cesa-Bianchi, and R.~Garnett (eds.), \emph{Advances in Neural Information Processing Systems}, volume~31. Curran Associates, Inc., 2018.

\bibitem[De~Brouwer et~al.(2019)De~Brouwer, Simm, Arany, and Moreau]{de2019gru}
Edward De~Brouwer, Jaak Simm, Adam Arany, and Yves Moreau.
\newblock Gru-ode-bayes: Continuous modeling of sporadically-observed time series.
\newblock \emph{Advances in neural information processing systems}, 32, 2019.

\bibitem[Dirac(1981)]{dirac1981principles}
Paul Adrien~Maurice Dirac.
\newblock \emph{The principles of quantum mechanics}.
\newblock Number~27. Oxford university press, 1981.

\bibitem[Gu et~al.(2021{\natexlab{a}})Gu, Goel, and R{\'e}]{gu2021efficiently}
Albert Gu, Karan Goel, and Christopher R{\'e}.
\newblock Efficiently modeling long sequences with structured state spaces.
\newblock \emph{arXiv preprint arXiv:2111.00396}, 2021{\natexlab{a}}.

\bibitem[Gu et~al.(2021{\natexlab{b}})Gu, Johnson, Goel, Saab, Dao, Rudra, and R{\'e}]{gu2021combining}
Albert Gu, Isys Johnson, Karan Goel, Khaled Saab, Tri Dao, Atri Rudra, and Christopher R{\'e}.
\newblock Combining recurrent, convolutional, and continuous-time models with linear state space layers.
\newblock \emph{Advances in neural information processing systems}, 34:\penalty0 572--585, 2021{\natexlab{b}}.

\bibitem[Jia \& Benson(2019)Jia and Benson]{jia2019neural}
Junteng Jia and Austin~R Benson.
\newblock Neural jump stochastic differential equations.
\newblock \emph{Advances in Neural Information Processing Systems}, 32, 2019.

\bibitem[Kiani et~al.(2022)Kiani, Balestriero, LeCun, and Lloyd]{kiani2022projunn}
Bobak Kiani, Randall Balestriero, Yann LeCun, and Seth Lloyd.
\newblock projunn: efficient method for training deep networks with unitary matrices.
\newblock In S.~Koyejo, S.~Mohamed, A.~Agarwal, D.~Belgrave, K.~Cho, and A.~Oh (eds.), \emph{Advances in Neural Information Processing Systems}, volume~35, pp.\  14448--14463. Curran Associates, Inc., 2022.

\bibitem[Kidger et~al.(2020)Kidger, Morrill, Foster, and Lyons]{NEURIPS2020_4a5876b4}
Patrick Kidger, James Morrill, James Foster, and Terry Lyons.
\newblock Neural controlled differential equations for irregular time series.
\newblock In H.~Larochelle, M.~Ranzato, R.~Hadsell, M.F. Balcan, and H.~Lin (eds.), \emph{Advances in Neural Information Processing Systems}, volume~33, pp.\  6696--6707. Curran Associates, Inc., 2020.

\bibitem[Kidger et~al.(2021{\natexlab{a}})Kidger, Foster, Li, and Lyons]{kidger2021neural}
Patrick Kidger, James Foster, Xuechen Li, and Terry~J Lyons.
\newblock Neural sdes as infinite-dimensional gans.
\newblock In \emph{International conference on machine learning}, pp.\  5453--5463. PMLR, 2021{\natexlab{a}}.

\bibitem[Kidger et~al.(2021{\natexlab{b}})Kidger, Foster, Li, and Lyons]{kidger2021efficient}
Patrick Kidger, James Foster, Xuechen~Chen Li, and Terry Lyons.
\newblock Efficient and accurate gradients for neural sdes.
\newblock \emph{Advances in Neural Information Processing Systems}, 34:\penalty0 18747--18761, 2021{\natexlab{b}}.

\bibitem[Lee et~al.(2022)Lee, Hasegawa, and Gao]{lee2022complex}
ChiYan Lee, Hideyuki Hasegawa, and Shangce Gao.
\newblock Complex-valued neural networks: A comprehensive survey.
\newblock \emph{IEEE/CAA Journal of Automatica Sinica}, 9\penalty0 (8):\penalty0 1406--1426, 2022.

\bibitem[Lezcano-Casado(2019)]{lezcano2019trivializations}
Mario Lezcano-Casado.
\newblock Trivializations for gradient-based optimization on manifolds.
\newblock In \emph{Advances in Neural Information Processing Systems, NeurIPS}, pp.\  9154--9164, 2019.

\bibitem[Lezcano-Casado \& Mart{\'i}nez-Rubio(2019)Lezcano-Casado and Mart{\'i}nez-Rubio]{lezcano2019cheap}
Mario Lezcano-Casado and David Mart{\'i}nez-Rubio.
\newblock Cheap orthogonal constraints in neural networks: A simple parametrization of the orthogonal and unitary group.
\newblock In \emph{International Conference on Machine Learning (ICML)}, pp.\  3794--3803, 2019.

\bibitem[Li et~al.(2021)Li, Gkoumas, Sordoni, Nie, and Melucci]{li2021quantum}
Qiuchi Li, Dimitris Gkoumas, Alessandro Sordoni, Jian-Yun Nie, and Massimo Melucci.
\newblock Quantum-inspired neural network for conversational emotion recognition.
\newblock In \emph{Proceedings of the AAAI Conference on Artificial Intelligence}, volume~35, pp.\  13270--13278, 2021.

\bibitem[Liu et~al.(2019)Liu, Xiao, Si, Cao, Kumar, and Hsieh]{liu2019neural}
Xuanqing Liu, Tesi Xiao, Si~Si, Qin Cao, Sanjiv Kumar, and Cho-Jui Hsieh.
\newblock Neural sde: Stabilizing neural ode networks with stochastic noise.
\newblock \emph{arXiv preprint arXiv:1906.02355}, 2019.

\bibitem[Morrill et~al.(2021{\natexlab{a}})Morrill, Kidger, Yang, and Lyons]{morrill2021neural}
James Morrill, Patrick Kidger, Lingyi Yang, and Terry Lyons.
\newblock Neural controlled differential equations for online prediction tasks.
\newblock \emph{arXiv preprint arXiv:2106.11028}, 2021{\natexlab{a}}.

\bibitem[Morrill et~al.(2021{\natexlab{b}})Morrill, Salvi, Kidger, and Foster]{pmlr-v139-morrill21b}
James Morrill, Cristopher Salvi, Patrick Kidger, and James Foster.
\newblock Neural rough differential equations for long time series.
\newblock In Marina Meila and Tong Zhang (eds.), \emph{Proceedings of the 38th International Conference on Machine Learning}, volume 139 of \emph{Proceedings of Machine Learning Research}, pp.\  7829--7838. PMLR, 18--24 Jul 2021{\natexlab{b}}.

\bibitem[Norcliffe et~al.(2021)Norcliffe, Bodnar, Day, Moss, and Li{\`o}]{norcliffe2021neural}
Alexander Norcliffe, Cristian Bodnar, Ben Day, Jacob Moss, and Pietro Li{\`o}.
\newblock Neural ode processes.
\newblock \emph{arXiv preprint arXiv:2103.12413}, 2021.

\bibitem[Oganesyan et~al.(2020)Oganesyan, Volokhova, and Vetrov]{oganesyan2020stochasticity}
Viktor Oganesyan, Alexandra Volokhova, and Dmitry Vetrov.
\newblock Stochasticity in neural odes: An empirical study.
\newblock \emph{arXiv preprint arXiv:2002.09779}, 2020.

\bibitem[Pal et~al.(2023)Pal, Zeng, Ravi, and Singh]{pal2023controlled}
Sourav Pal, Zhanpeng Zeng, Sathya~N Ravi, and Vikas Singh.
\newblock Controlled differential equations on long sequences via non-standard wavelets.
\newblock 2023.

\bibitem[Rubanova et~al.(2019)Rubanova, Chen, and Duvenaud]{NEURIPS2019_42a6845a}
Yulia Rubanova, Ricky T.~Q. Chen, and David~K Duvenaud.
\newblock Latent ordinary differential equations for irregularly-sampled time series.
\newblock In H.~Wallach, H.~Larochelle, A.~Beygelzimer, F.~d\textquotesingle Alch\'{e}-Buc, E.~Fox, and R.~Garnett (eds.), \emph{Advances in Neural Information Processing Systems}, volume~32. Curran Associates, Inc., 2019.

\bibitem[Rusch \& Mishra(2021)Rusch and Mishra]{rusch2021unicornn}
T~Konstantin Rusch and Siddhartha Mishra.
\newblock Unicornn: A recurrent model for learning very long time dependencies.
\newblock In \emph{International Conference on Machine Learning}, pp.\  9168--9178. PMLR, 2021.

\bibitem[Salvi et~al.(2022)Salvi, Lemercier, and Gerasimovics]{salvi2022neural}
Cristopher Salvi, Maud Lemercier, and Andris Gerasimovics.
\newblock Neural stochastic pdes: Resolution-invariant learning of continuous spatiotemporal dynamics.
\newblock \emph{Advances in Neural Information Processing Systems}, 35:\penalty0 1333--1344, 2022.

\bibitem[Schirmer et~al.(2022)Schirmer, Eltayeb, Lessmann, and Rudolph]{schirmer2022modeling}
Mona Schirmer, Mazin Eltayeb, Stefan Lessmann, and Maja Rudolph.
\newblock Modeling irregular time series with continuous recurrent units.
\newblock In \emph{International Conference on Machine Learning}, pp.\  19388--19405. PMLR, 2022.

\bibitem[Shafran et~al.(2019)Shafran, Bagby, and Skerry-Ryan]{shafran2019complex}
Izhak Shafran, Tom Bagby, and R.~J. Skerry-Ryan.
\newblock Complex evolution recurrent neural networks (cernns).
\newblock 2019.

\bibitem[Smith et~al.(2022)Smith, Warrington, and Linderman]{smith2022simplified}
Jimmy~TH Smith, Andrew Warrington, and Scott~W Linderman.
\newblock Simplified state space layers for sequence modeling.
\newblock \emph{arXiv preprint arXiv:2208.04933}, 2022.

\bibitem[Voigtlaender(2023)]{VOIGTLAENDER202333}
Felix Voigtlaender.
\newblock The universal approximation theorem for complex-valued neural networks.
\newblock \emph{Applied and Computational Harmonic Analysis}, 64:\penalty0 33--61, 2023.
\newblock ISSN 1063-5203.
\newblock \doi{https://doi.org/10.1016/j.acha.2022.12.002}.

\bibitem[Xia et~al.(2021)Xia, Suliafu, Ji, Nguyen, Bertozzi, Osher, and Wang]{xia2021heavy}
Hedi Xia, Vai Suliafu, Hangjie Ji, Tan Nguyen, Andrea Bertozzi, Stanley Osher, and Bao Wang.
\newblock Heavy ball neural ordinary differential equations.
\newblock \emph{Advances in Neural Information Processing Systems}, 34:\penalty0 18646--18659, 2021.

\end{thebibliography}
\bibliographystyle{iclr2023_conference_tinypaper}

\appendix
\section{Model architecture}\label{app:architecture}

In this Appendix, we give details about the structure of the four NQDE models (for the vector field of the controlled differential equation).

Using ProjUNN, the model for the first architecture where we combine the representation of complex values (for each class) before the final linear layer is seen below. This model has 5052 trainable model parameters.

\begin{code}
NeuralQDE(
  (func): QDEFunc(
    (linear1): Linear(in\_features=4, out_features=32, bias=True)
    (linear2): Linear(in\_features=64, out_features=12, bias=True)
    (rnn\_layer): OrthogonalRNN(
      (recurrent_kernel): Linear(in\_features=32, out\_features=32, bias=False)
      (input_kernel): Linear(in\_features=32, out_features=32, bias=False)
      (nonlinearity): ReLU()
    )
    (rnn_layer2): OrthogonalRNN(
      (recurrent_kernel): Linear(in_features=32, out_features=32, bias=False)
      (input_kernel): Linear(in_features=32, out_features=32, bias=False)
      (nonlinearity): ReLU()
    )
  )
  (initial): Linear(in_features=3, out_features=4, bias=True)
)
\end{code}

Using ProjUNN, the model for the second architecture where we combine the representation of complex values (for each class) after the final linear layer is seen below.  This model has 4470 trainable model parameters.

\begin{code}
NeuralQDE(
  (func): QDEFunc2(
    (linear1): Linear(in_features=4, out_features=32, bias=True)
    (linear2): Linear(in_features=32, out_features=6, bias=True)
    (rnn_layer): OrthogonalRNN(
      (recurrent_kernel): Linear(in_features=32, out_features=32, bias=False)
      (input_kernel): Linear(in_features=32, out_features=32, bias=False)
      (nonlinearity): ReLU()
    )
    (rnn_layer2): OrthogonalRNN(
      (recurrent_kernel): Linear(in_features=32, out_features=32, bias=False)
      (input_kernel): Linear(in_features=32, out_features=32, bias=False)
      (nonlinearity): ReLU()
    )
  )
  (initial): Linear(in_features=3, out_features=4, bias=True)
)
)    
\end{code}

Using GeoTorch, the model for the third architecture where we combine the representation of complex values (for each class) before the final linear layer is seen below. This model has 3068 trainable model parameters.

\begin{code}
NeuralQDE(
  (func): QDEFunc3(
    (linear1): Linear(in_features=4, out_features=32, bias=True)
    (linear2): ParametrizedLinear(
      in_features=32, out_features=32, bias=True
      (parametrizations): ModuleDict(
        (weight): ParametrizationList(
          (0): Stiefel(n=32, k=32, triv=linalg_matrix_exp)
        )
      )
    )
    (linear3): ParametrizedLinear(
      in_features=32, out_features=32, bias=True
      (parametrizations): ModuleDict(
        (weight): ParametrizationList(
          (0): Stiefel(n=32, k=32, triv=linalg_matrix_exp)
        )
      )
    )
    (linear4): Linear(in_features=64, out_features=12, bias=True)
  )
  (initial): Linear(in_features=3, out_features=4, bias=True)
)
\end{code}

Using GeoTorch, the model for the fourth architecture where we combine the representation of complex values (for each class) after the final linear layer is seen below. This model has 2486 trainable model parameters.

\begin{code}
NeuralQDE(
  (func): QDEFunc4(
    (linear1): Linear(in_features=4, out_features=32, bias=True)
    (linear2): ParametrizedLinear(
      in_features=32, out_features=32, bias=True
      (parametrizations): ModuleDict(
        (weight): ParametrizationList(
          (0): Stiefel(n=32, k=32, triv=linalg_matrix_exp)
        )
      )
    )
    (linear3): ParametrizedLinear(
      in_features=32, out_features=32, bias=True
      (parametrizations): ModuleDict(
        (weight): ParametrizationList(
          (0): Stiefel(n=32, k=32, triv=linalg_matrix_exp)
        )
      )
    )
    (linear4): Linear(in_features=32, out_features=6, bias=True)
  )
  (initial): Linear(in_features=3, out_features=4, bias=True)
)    
\end{code}

\section{Hyper-parameter settings} \label{app: hyperparams}
The hyperparamter of the experiments can be seen in Table~\ref{tab:hyper}. 

\begin{table}[h]
\begin{center}
\begin{tabular}{c|c|c|c}
\hline
\hline
\textbf{model}            & \textbf{epoch}      & \textbf{lr} & \textbf{lin\_size}  \vspace{0.2em}   \\

\hline
\hline
NQDE1\_unn      & 20 & 0.002    & 32  \\
NQDE2\_unn      & 20  & 0.002    & 32  \\
NQDE3\_geo & 20 & 0.001    & 32   \\
NQDE4\_geo & 20 & 0.001     & 32  
\end{tabular}\caption{Hyperparameter settings for the four models}\label{tab:hyper}
\end{center}
\end{table}

\section{Related neural differential equation work}\label{app:related_work}

Seen as a continuous time generalisation of discrete models, differential equation based neural networks enjoyed successes in recent years.
\citet{NEURIPS2018_69386f6b} proposed neural ordinary differential equations (ODEs) and demonstrated its performance on classification and time-series generation problems. In a follow-up work \citet{NEURIPS2019_42a6845a} uses a neural ODE to model the (autonomous) evolution of the hidden state before the next observation arrives, and obtained encouraging results. This approach is then generalised to the neural controlled differential equations (CDEs) \citep{NEURIPS2020_4a5876b4}. \citet{kidger2021neural} also worked on an extension to neural stochastic differential equations (SDEs) and application to GANs for learning path distribution. Neural rough differential equations (RDEs) are proposed as an improvement on neural CDEs by summarising sub-intervals using log-signatures, deriving the theoretical justifications from the log-ODE method \citep{pmlr-v139-morrill21b}. \citet{pal2023controlled} proposed another way of adapting neural CDEs to handle long time series. 

Heavy ball method is proposed to improve training of neural ODE in \citep{xia2021heavy}. 
\citet{norcliffe2021neural} combines neural ODEs with the so-called neural processes so that the model (the trained neural network) can adapted to incoming data stream. 
\citet{morrill2021neural} uses neural CDEs for online prediction task.
\citet{jia2019neural} extends neural ODEs to handle stochastic jumps and discussed training techniques when the latent state has discontinuities.
\citet{oganesyan2020stochasticity} and \citet{ liu2019neural} view neural SDEs as a stochastic regularisation technique for training neural ODEs and evaluated empirically the performances. \citet{kidger2021efficient} proposed improved technique for training of neural SDEs. \citet{salvi2022neural} proposed using neural network to learn the dynamics of stochastic partial differential equations (SPDEs) and showed that for several well-known SPDE dynamics, the solver can be learned faster than traditional numerical solvers of SPDE. Deep state space models such as the S4 \citep{gu2021efficiently, gu2021combining} can effectively model long range dependency in sequence modelling,
this is subsequently improved by \citet{smith2022simplified}. 

The survey \citep{lee2022complex} discuses split-complex neural networks, which split the complex value input into real and imaginary parts which are fed into a real-valued neural network, that could have real-valued weight and real activation or complex-valued weights and real activation. Some training instability issues are highlighted.
There are few applications of complex RNNs. 
\citet{de2019gru} improves the variational autoencoder application of neural ODE by combining neural ODE with a continuous version of a GRU. \citet{schirmer2022modeling} uses an SDE with a Kalman filter to connect observations at different timestamps in an RNN. \citet{rusch2021unicornn} studied a restricted class of ODE discretised RNN based on Hamiltonian system ODE.
\citet{li2021quantum} implicitly made a connection between unitary RNNs and quantum-inspired theory, and experimented a unitary RNN on an emotional recognition problem from multi-modal time-series data. 

In terms of function approximation power,
\citet{VOIGTLAENDER202333}
finds that unlike the classical case of real networks, the set of “good activation functions”—which give rise to networks with the universal approximation property—differs significantly depending on whether one considers deep networks or shallow networks. 
For deep networks with at least two hidden layers, the universal approximation property holds as long as \(\sigma\), the activation function, is neither a polynomial, a holomorphic function, nor an antiholomorphic function. Shallow networks, on the other hand, are universal if and only if the real part or the imaginary part of \(\sigma\) is not a polyharmonic function.

\end{document}